# Slice Sampling Particle Belief Propagation

Oliver Müller, Michael Ying Yang, and Bodo Rosenhahn
Institute for Information Processing (TNT), Leibniz University Hannover, Germany
{omueller, yang, rosenhahn}@tnt.uni-hannover.de

## Abstract

*Inference in continuous label Markov random fields is a challenging task. We use particle belief propagation (PBP) for solving the inference problem in continuous label space. Sampling particles from the belief distribution is typically done by using Metropolis-Hastings (MH) Markov chain Monte Carlo (MCMC) methods which involves sampling from a proposal distribution. This proposal distribution has to be carefully designed depending on the particular model and input data to achieve fast convergence. We propose to avoid dependence on a proposal distribution by introducing a slice sampling based PBP algorithm. The proposed approach shows superior convergence performance on an image denoising toy example. Our findings are validated on a challenging relational 2D feature tracking application.*

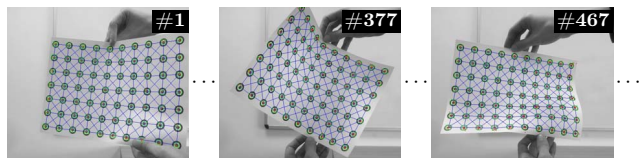

Figure 1. Relational 2D feature tracking example.

## 1. Introduction

Markov Random Fields (MRFs) are a powerful tool for modeling relational dependencies among observations. Inference in such models is an inherent problem which has been widely addressed in the past. MRFs, and hence its inference methods, can be classified in two categories: discretely and continuously labeled problems. Numerous optimization approaches for discrete labels have been proposed, from binary labeled Graph Cuts [4], to multi-label tree reweighted message passing [17, 7]. In this paper, we deal with continuous labeled MRFs where we use a particle belief propagation (PBP) approach [6]. The efficiency of such particle based approaches highly depends on the sampling scheme used to explore the label space. Previous approaches use Metropolis-Hastings (MH) Markov chain Monte Carlo (MCMC) methods for particle sampling. The performance of these methods depends on a carefully designed proposal distribution.

**Contributions.** We propose a novel sampling technique for PBP based on slice sampling [12]. This method exploits the structure of the PBP message passing equations for direct sampling from the target distribution and does not depend on a proposal distribution which is difficult to tune. We show the superiority of our method theoretically on a simplified toy application on image denoising. Our findings are then verified on a complex 2D relational feature tracking application as shown in Fig. 1. We furthermore provide a publicly available database of image sequences for feature tracking applications including manually labeled groundtruth data [11].

The rest of the paper is organized as follows. Section 2 provides an overview of related work. Section 3 introduces notations and definitions used throughout the paper and gives a short introduction to slice sampling. Our proposed approach is described in detail in Sect. 4. In Sect. 5 we present a thorough evaluation of our method compared to the state-of-the-art and propose a 2D relational feature tracking application. We conclude our findings in Sect. 6.

## 2. Related Work

Most works on MRF optimization specialize on a discrete label space [4, 17, 7]. Often such approaches are hard to apply on tasks where a continuous label space would be a more natural choice, such as feature tracking with relational constraints [14, 9].

Loopy belief propagation is a prominent method using a local message passing mechanism for coordinating the optimal labeling of neighboring nodes. These methods work on discrete label spaces. The computational complexity is $\mathcal{O}(n^2)$ over the number of discrete labels $n$, making computations with many labels for approximating near-continuous models intractable [16].

Recently, message passing approaches working in continuous rather than discrete label space were proposed

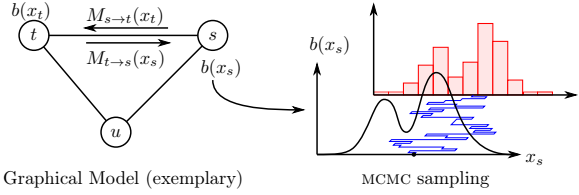

Figure 2. Particle Belief Propagation framework. Left: Message passing mechanism. Right: MCMC particle sampling of the belief $b(x_s)$ with an exemplary MCMC sampling chain of one particle (blue) and its corresponding histogram (red).

[6, 8, 13, 16]. These approaches use MCMC methods to approximate the message distributions. To the best of our knowledge, all previously proposed MCMC based belief propagation methods use Metropolis-Hastings (MH) sampling. This sampling strategy consists of two steps: (a) sampling a candidate particle from an easy to sample *proposal distribution*, and (b) accept or reject the candidate depending on a transition probability [18]. Applying this sampling technique involves a careful design of the proposal distribution, which is a compromise between exploring the label space (using a broad proposal distribution) and maximizing the transition acceptance ratio (minimize sample moves) at the same time.

Throughout the paper we show that considering alternative sampling techniques can be advantageous. We propose to use slice sampling [12] instead of MH, rendering proposal distribution selection obsolete in the context of PBP.

To demonstrate superior performance of our method on a real world problem we propose a relational feature tracking application inspired by [9, 14] in the experiment section. Some related works such as [15, 5] propose to formulate feature tracking as a discrete labeling problem and use global optimization algorithms (i.e. linear programming or dynamic programming). Such approaches need some sort of label pruning in order to keep computational complexity low. Closely related methods use belief propagation combined with particle filtering [19, 9, 14], but still use proposal distributions for particle perturbation which introduces sensible optimization parameter tuning.

## 3. Definitions and Notation

### 3.1. Markov Random Field

Let $\mathcal{V}$ be a set of nodes and $\mathcal{N}_s \subset \mathcal{V}$ the set of neighboring nodes to $s \in \mathcal{V}$. For every node $s$ there is a label $x_s$ from the *label space* $\mathcal{L}_s$. The product $\mathcal{L} = \prod_{s \in \mathcal{V}} \mathcal{L}_s$ is the space of configurations $\mathbf{x} = \{x_s\}_{s \in \mathcal{V}}$. A Markov random field potential energy is given by:

$$E(\mathbf{x}) = \sum_{s \in \mathcal{V}} \psi_s(x_s) + \sum_{s \in \mathcal{V}} \sum_{t \in \mathcal{N}_s} \psi_{s,t}(x_s, x_t) \quad (1)$$

with a *unary potential* function $\psi_s(x_s)$ and a *binary potential* function $\psi_{s,t}(x_s, x_t)$. Then $p(\mathbf{x}) \propto \exp[-E(\mathbf{x})]$ defines a Markov random field (MRF).

We consider the problem of computing the maximum marginals: $\mu(x_s) = \max_{\mathbf{x}'|x'_s=x_s} p(\mathbf{x}')$[1].

### 3.2. Max-Product Particle Belief Propagation

In the following we summarize the max-product particle belief propagation algorithm [8, 3]. The energy term $E(\mathbf{x})$ is approximated by particles such that the label space $\mathcal{L}_s$ of each node $s$ in the MRF is represented by a set of particles $P_s = \{x_s^{(1)}, \ldots, x_s^{(p)}\}$, where $p$ is the number of particles per node. Then the estimated *belief* $b_s^n(x_s^{(i)})$ or *log disbelief* $B_s^n(x_s^{(i)}) = -\log(b_s^n(x_s^{(i)}))$ of node $s$ at iteration $n$ is calculated as follows [3]:

$$B_s^n(x_s^{(i)}) = \psi_s(x_s^{(i)}) + \sum_{t \in \mathcal{N}_s} M_{t \to s}^n(x_s^{(i)}), \quad (2)$$

where the *messages* $M_{t \to s}^n(x_s)$ for $x_s \in P_s$ from node $t$ to node $s$ are:

$$M_{t \to s}^n(x_s) = \min_{x_t \in P_t}[\psi_{s,t}(x_s, x_t) + B_t^{n-1}(x_t) - M_{s \to t}^{n-1}(x_t)]. \quad (3)$$

Note that the log disbelief $B_s^n(x_s)$ and the messages $M_{t \to s}^n(x_s)$ can be calculated for all continuous values $x_s \in \mathcal{L}_s$ rather than only on the particle set $P_s$. On the other hand, the messages from node $s$ to node $t$ are approximated only using the particles $x_t$ from the particle set $P_t = \{x_t^{(1)}, \ldots, x_t^{(p)}\}$ of node $t$.

Messages and log disbeliefs are calculated iteratively for $n = 1, \ldots, N$ iterations. An estimate of the most likely configuration can be obtained with

$$\hat{x}_s = \arg \min_{x_s} B_s^N(x_s). \quad (4)$$

The main issue in PBP lies in how to sample new particles $x_s^n \sim B_s^n(x_s)$. Typically, the Metropolis-Hastings (MH) MCMC method is used. This method requires a proposal distribution $q$ where new particles can be easily sampled from. Typically a Gaussian function $q = p_\sigma$ with a predefined standard deviation $\sigma$ is used.

Figure 2 shows a schematic overview of the PBP framework. Algorithm 1 summarizes the Metropolis-Hastings based max-product particle belief propagation algorithm (MH-PBP).

Typically, $q$ needs to be carefully adjusted to the true belief distribution. This introduces a dependency on prior knowledge about how the labels are distributed in the label space. In the following we propose to replace the MH sampling step by a slice sampling approach which does not depend on proposal distribution selection.

---
[1]Backtracking can be used to compute the MAP-configuration $\mathbf{x}^* = \arg \max_{\mathbf{x}} p(\mathbf{x})$ from the max-marginals [8].

## Algorithm 1 MH-PBP [8, 3]

**Input:** Initial set of particles: $\{x_s^{(i)}\}_{i=1,\ldots,p}$, proposal distribution $p_\sigma$

1: Initialize the messages $M_{t\to s}^0(x_s)$ and log disbelief $B_s^0(x_s^{(i)})$ with zero $\forall s, t$
2: **for** BP iteration $n = 1$ to $N$ **do**
3:    **for** each node $s$ and each particle $i = 1, \ldots, p$ **do**
4:       Initialize sampling chain $x_s^{(i)\langle 0 \rangle} \leftarrow x_s^{(i)}$
5:       **for** MCMC iteration $m = 1, \ldots, M$ **do**
6:          Sample $\bar{x}_s^{(i)\langle m \rangle} \sim p_\sigma(x \mid x_s^{(i)\langle m-1 \rangle})$ from proposal distribution $p_\sigma$
7:          Calc. belief $B_s^n(\bar{x}_s^{(i)\langle m \rangle})$ from Eqs. (2), (3)
8:          Sample $u \sim \mathcal{U}_{[0,1]}(u)$
9:          **if** $B_s^n(\bar{x}_s^{(i)\langle m \rangle}) < B_s^n(x_s^{(i)\langle m \rangle}) - \log(u)$ **then**
10:            Accept: $x_s^{(i)\langle m \rangle} \leftarrow \bar{x}_s^{(i)\langle m \rangle}$
11:          **end if**
12:       **end for**
13:       $x_s^{(i)} \leftarrow x_s^{(i)\langle M \rangle}$
14:    **end for**
15:    Normalize messages and beliefs
16: **end for**

### 3.3. MCMC Slice Sampling

In this section we briefly summarize the concept of slice sampling [12, 1] which is defined in a general MCMC sampling framework. Suppose we are given a distribution $q(x)$ and want to sample from this distribution, *i.e.* MCMC sampling of $M$ samples $x^{\langle 1 \rangle}, x^{\langle 2 \rangle}, \ldots, x^{\langle M \rangle}$:

$$x^{\langle m \rangle} \sim q(x \mid x^{\langle m-1 \rangle}), \quad (5)$$

given an initial sample $x^{\langle 0 \rangle}$.

Note that in the PBP framework, there is a MCMC sampling chain $\{x_s^{(i)\langle m \rangle}\}_{m=1,\ldots,M}$ for each particle $x_s^{(i)}$. MCMC sampling could be done using several sampling techniques such as Metropolis-Hastings (MH) or Gibbs sampling (provided the conditional distributions are easy to sample from). Metropolis-Hastings sampling has the drawback of requiring a proposal distribution. Choosing the proposal distribution is very often a difficult task and introduces a compromise between reducing the rejection rate and obtaining large random moves [1].

In slice sampling, an auxiliary variable $u \in \mathbb{R}$ is introduced and the target distribution $q(x)$ is extended to

$$q^*(x, u) = \begin{cases} 1 & \text{if } u \in [0, q(x)] \\ 0 & \text{otherwise} \end{cases} \quad (6)$$

Sampling is then done by uniformly drawing the auxiliary variable $u$ (defining the *slice*) and given this, uniformly drawing the new sample from an interval $A$ defined over $u$ as follows:

$$u^{\langle m \rangle} \sim q(u \mid x^{\langle m-1 \rangle}) = \mathcal{U}_{[0, q(x^{\langle m-1 \rangle})]}(u) \quad (7)$$

$$x^{\langle m \rangle} \sim q(x \mid u^{\langle m \rangle}) = \mathcal{U}_A(x), \quad (8)$$

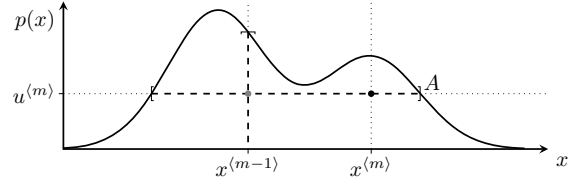

Figure 3. Slice Sampling [12, 1]

where $\mathcal{U}_I$ is the uniform distribution over an interval $I$ and $A = \{x; q(x) \geq u^{\langle m \rangle}\}$. Figure 3 shows an exemplary slice sampling step.

Assume that $q(x)$ can be decomposed in $L$ functions $f_l(x)$ such that $q(x) \propto \prod_{l=1}^L f_l(x)$. Then we can sample over $q(x)$ by introducing $L$ auxiliary variables $u_1, \ldots, u_L$:

$$u_1^{\langle m \rangle} \sim q(u_1 \mid x^{\langle m-1 \rangle}) = \mathcal{U}_{[0, f_1(x^{\langle m-1 \rangle})]}(u_1) \quad (9)$$

$$\vdots$$

$$u_L^{\langle m \rangle} \sim q(u_L \mid x^{\langle m-1 \rangle}) = \mathcal{U}_{[0, f_L(x^{\langle m-1 \rangle})]}(u_L) \quad (10)$$

$$x^{\langle m \rangle} \sim q(x \mid u_1^{\langle m \rangle}, \ldots, u_L^{\langle m \rangle}) = \mathcal{U}_{A^{\langle m \rangle}}(x), \quad (11)$$

where $A^{\langle m \rangle} = \{x; f_l(x) \geq u_l^{\langle m \rangle}, l = 1, \ldots, L\}$ [1].

The main difficulty lies in determining the interval $A$. Fortunately it turns out, that in the max-product particle belief propagation framework the sampling interval $A$ can be determined efficiently as shown in the following section.

## 4. Slice Sampling Particle Belief Propagation

Our main contribution is presented in this section. We propose to sample particles from the belief $b(x_s)$ using slice sampling rather than Metropolis-Hastings sampling. For applying the slice sampler, the sampling interval $A^{(i)\langle m \rangle}$ needs to be determined for the $i$th particle of node $s$ and for the $m$th MCMC iteration which we can uniformly sample the particle $x_s^{(i)\langle m \rangle}$ from. The superscripts $(i)\langle m \rangle$ are omitted in the following for better readability.

The goal is to determine the sampling interval $A$. Given the potential functions $\psi_s(x_s)$ and $\psi_{s,t}(x_s, x_t)$, it is assumed that the intervals

$$A_{\psi_s}(\bar{u}) = \{x_s; \psi_s(x_s) \leq \bar{u}\} \quad \text{and} \quad (12)$$

$$A_{\psi_{s,t}}^{x_t}(\bar{u}) = \{x_s; \psi_{s,t}(x_s, x_t) \leq \bar{u}\} \quad (13)$$

can be computed analytically. Note that computations are done in negative log space, thus a slice interval $\{x; f(x) \geq u\}$ is transformed to $\{x; -\log(f(x)) \leq \bar{u}\}$, where $\bar{u}$ is the negative logarithm of a uniformly sampled value.

The final sampling interval $A$ can be computed from these intervals as shown below. If the intervals cannot be computed analytically then an approximated interval $\tilde{A}$ may be still computed and rejection sampling can be applied [1].

1131

**Algorithm 2** S-PBP

**Input:** Initial set of particles: $\{x_s^{(i)}\}_{i=1,\ldots,p}$
1: Initialize the messages $M_{t \to s}^0(x_s)$ and log disbelief $B_s^0(x_s^{(i)})$ with zero $\forall s, t$
2: **for** BP iteration $n = 1$ to $N$ **do**
3:   **for** each node $s$ and each particle $i = 1, \ldots, p$ **do**
4:     Initialize sampling chain $x_s^{(i)\langle 0 \rangle} \leftarrow x_s^{(i)}$
5:     **for** MCMC iteration $m = 1, \ldots, M$ **do**
6:       Sample $\bar{u}_l = F_l(x_s^{(i)\langle m-1 \rangle}) - \log(u_l)$ where $u_l \sim \mathcal{U}_{[0,1]}(u)$ for $l = 0, \ldots, |\mathcal{N}_s|$
7:       Compute $A^{(i)\langle m \rangle}$ from Eqs. (15), (16), (17)
8:       Sample $\bar{x}_s^{(i)\langle m \rangle} \sim \mathcal{U}_{A^{(i)\langle m \rangle}}(x)$
9:       Calc. belief $B_s^n(\bar{x}_s^{(i)\langle m \rangle})$ from Eqs. (2), (3)
10:       **if** $F_l(\bar{x}_s^{(i)\langle m \rangle}) \leq \bar{u}_l$ for $l = 0, \ldots, |\mathcal{N}_s|$ **then**
11:         Accept: $x_s^{(i)\langle m \rangle} \leftarrow \bar{x}_s^{(i)\langle m \rangle}$
12:       **end if**
13:     **end for**
14:     $x_s^{(i)} \leftarrow x_s^{(i)\langle M \rangle}$
15:   **end for**
16:   Normalize messages and beliefs
17: **end for**

The log disbelief can be decomposed as follows:

$$B(x_s) = \sum_{l=0}^{|\mathcal{N}_s|} F_l(x_s) \quad (14)$$

with $F_0(x_s) = \psi_s(x_s)$ and $F_j(x_s) = M_{t^{(j)} \to s}(x_s)$ where $t^{(j)}$ is the $j$-th neighbor of $s$. From this follows the decomposition of the sampling interval

$$A = \bigcap_{l=0}^{|\mathcal{N}_s|} A_l, \quad \text{with} \quad A_l = \{x \, ; \, F_l(x) \leq \bar{u}_l\}. \quad (15)$$

Using the definitions of Eqs. (2, 3), we obtain for $A_l$:

$$A_0 = A_{\psi_s}(\bar{u}_1) \quad (16)$$

$$\begin{aligned}
A_j &= \{x_s \, ; \, M_{t^{(j)} \to s}(x_s) \leq \bar{u}_j\} \\
&= \{x_s \, ; \, \min_{x_t \in P_t} G_j^{x_t}(x_s) \leq \bar{u}_j\} \\
&= \bigcup_{x_t \in P_t} \{x_s \, ; \, G_j^{x_t}(x_s) \leq \bar{u}_j\} \\
&= \bigcup_{x_t \in P_t} A_{\psi_{s,t}}^{x_t}(\bar{u}_j - B_t(x_t) + M_{s \to t}(x_t)), \quad (17)
\end{aligned}$$

where $G_j^{x_t}(x_s) = \psi_{s,t}(x_s, x_t) + B_t^{n-1}(x_t) - M_{s \to t}^{n-1}(x_t)$. This result shows that $A$ only depends on the given intervals $A_{\psi_s}(\bar{u})$ and $A_{\psi_{s,t}}^{x_t}(\bar{u})$ which are defined by the unary and binary potential functions $\psi_s$ and $\psi_{s,t}$. Algorithm 2 summarizes the proposed method.

We further refer to the proposed technique as S-PBP (slice sampling particle belief propagation).

**Example.** Consider a quadratic unary potential function $\phi_s(x_s) = (x_s - d_s)^2$. Then $A_{\phi_s}(\bar{u})$ has the closed form solution $A_{\phi_s}(\bar{u}) = \{x_s : (x_s - d_s)^2 \leq \bar{u}\} = [d_s - \sqrt{\bar{u}}, d_s + \sqrt{\bar{u}}]$. Similarly, the closed form solution for $\phi_{s,t}(x_s, x_t) = (x_s - x_t)^2$ is $A_{\phi_{s,t}}^{x_t}(\bar{u}) = [x_t - \sqrt{\bar{u}}, x_t + \sqrt{\bar{u}}]$.

**Multidimensional Bounds.** In order to deal with multidimensional label spaces, i.e. $\mathcal{L}_s \in \mathbb{R}^d$ for $d > 1$, we propose to randomly sample one dimension in each MCMC step and slice sample on this dimension while the other dimensions are held fixed.

**Analytic Bounds Calculation.** Assume the unary and/or binary potential functions $\psi_s$ and $\psi_{st}$ are given as an analytic function. Then one can use standard computer algebra solvers for defining $A_{\psi_s}(u)$ and/or $A_{\psi_{s,t}}^{x_t}(u)$. We have implemented our S-PBP framework in MATLAB® with MEX and use the MATLAB®-MUPAD® interface to solve the inequalities automatically. This way no manual work has to be done.

## 5. Experiments

### 5.1. Image Denoising

For analyzing the random walk behaviour of our method we have chosen the application of image denoising due to its relatively simple model structure. The basic image denoising model is as follows:

$$\begin{aligned}
\psi_s(x_s) &= \theta_1 (x_s - d_s)^2, \\
\psi_{s,t}(x_s, x_t) &= \theta_2 \min\{\theta_3, (x_s - x_t)^2\}.
\end{aligned} \quad (18)$$

For minimizing particle noise in the final estimation result an annealing scheme is used where the target belief distribution is modified to $b_s^n(x_s^{(i)})^{1/T_n}$, where $T_n = T_0 \cdot (T_N/T_0)^{n/N}$ is the temperature at PBP iteration $n$, $T_0$ is the start temperature, and $T_N$ the end temperature. Given this annealing scheme the temperature is successively reduced for each new iteration $n$.

The evaluation was done on an example image as shown in Fig. 4. The training and testing sets each include 10 noisy image instances with Gaussian noise standard deviation $\sigma = 0.05$ (where image intensity $\in [0, 1]$). Training of the parameter vector $\theta = \{\theta_1, \theta_2, \theta_3\}$ is done by minimizing the *empirical risk* $R(\theta) = \frac{1}{K} \sum_{i=1}^{K} L(\mathbf{x}_\theta^{(i)}, \mathbf{y}^{(i)})$ given the *loss function* $L(\mathbf{x}, \mathbf{y}) = \|\mathbf{x} - \mathbf{y}\|_2^2$ where $\{\mathbf{y}^{(i)}, \mathbf{d}^{(i)}\}$ is the training data pair with groundtruth $\mathbf{y}^{(i)}$ and noisy observation $\mathbf{d}^{(i)}$. $\mathbf{x}_\theta^{(i)}$ is the MAP estimate given $\mathbf{d}^{(i)}$ and the parameter $\theta$. Learned parameters are $\theta_1 = 0.756$, $\theta_2 = 1.170$, $\theta_3 = 0.0059$.

**Comparing S-PBP with MH-PBP.** We further compared the efficiency of the slice sampling method to the Metropolis-Hastings sampling applied on the image denoising problem. For the experimental setup we use $N = 100$ PBP iterations, $p = 5$ particles, and a temperature schedule of $T_0 = 1$ to $T_N = 10^{-4}$. An MCMC chain of $M = 500$ samples is generated for each particle and in each PBP iteration. The iteration numbers are chosen to be more than sufficiently large in order to guarantee convergence and to collect statistical information in the MCMC chains in steady-state situations. For the MH-PBP proposal distribution the family of Gaussian distributions $p_\sigma(x \mid x^{\langle m-1 \rangle}) = (2\pi\sigma^2)^{-0.5} \cdot \exp[-0.5(x - x^{\langle m-1 \rangle})^2 \cdot \sigma^{-2}]$ is used. In order to provide a fair comparison the proposal distribution is adapted to the current temperature by using $p_\sigma(x \mid x^{\langle m-1 \rangle})^{1/T_n}$ instead.

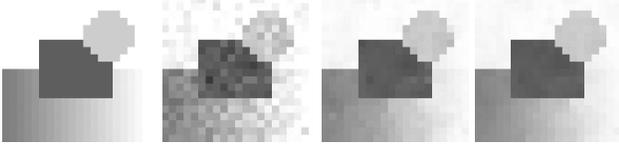

Figure 4. Denoising example: Groundtruth (left), noisy input example (middle left), reconstruction with MH-PBP (middle right), reconstruction with our proposed S-PBP method (right).

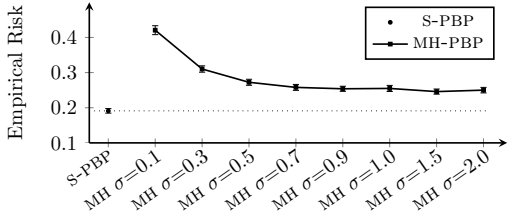

Figure 5. Comparison of the empirical risk for S-PBP and MH-PBP with different proposal distributions.

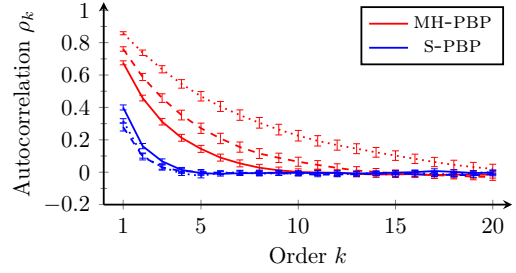

Figure 6. Comparison of S-PBP and MH-PBP at different PBP iterations (dotted $n = 30$, dashed $n = 50$, and solid $n = 70$) using an annealing schedule.

Figure 5 shows a comparison of the empirical risk for different MH-PBP proposal distributions. For $\sigma > 0.7$ the empirical risk stays nearly at the same level and thus we selected $\sigma = 0.7$ for further experiments. Another observation is that S-PBP outperforms MH-PBP in terms of minimal empirical risk. This is because the reconstructed images with MH-PBP have always much higher noise than images reconstructed with S-PBP. This effect can be significantly reduced by averaging over particles instead of only selecting the best one as stated in Eq. (4).

For comparing the random walk behavior of the MCMC sampling chains from S-PBP and MH-PBP, the normalized autocorrelation function

$$\rho_k = \frac{\sum_{m=1}^{M-k}(x^{\langle m \rangle} - \bar{x})(x^{\langle m-k \rangle} - \bar{x})}{\sum_{m=1}^{M-k}(x^{\langle m \rangle} - \bar{x})^2}, \quad (19)$$

where $\bar{x} = \frac{1}{M}\sum x^{\langle m \rangle}$, is used [18]. Only the last 50 % of the MCMC chain is considered to skip any burn-in phase. Figure 6 shows a comparison of the first 20 $k$th order autocorrelation of S-PBP and MH-PBP at different PBP iterations $n$ (and thus at different temperatures $T_n$). It can be observed that the MH-PBP method produces a much higher autocorrelation than the S-PBP method, thus the MCMC chain mixing behaviour of S-PBP outperforms MH-PBP.

### 5.2. Relational Feature Tracking

We propose to apply our S-PBP algorithm on a 2D relational feature tracking system inspired by [9, 14] as a more complex application.

#### 5.2.1 Tracker Model

The proposed feature tracker uses a pairwise MRF model. The model is separated into two parts: (a) the unary potentials are derived from a feature patch matching model, and (b) the binary potentials encode the relative positioning of the features to each other. The label space of the MRF is the space of feature poses including the local central patch position, patch rotation, and scale. The proposed MRF model is as follows:

$$E(\mathbf{x}) = \sum_{s \in \mathcal{V}} \psi_s(x_s) + \sum_{s \in \mathcal{V}} \sum_{t \in \mathcal{N}_s} \alpha \cdot \psi_{s,t}(x_s, x_t), \quad (20)$$

where the unary potential function

$$\psi_s(x_s) = \chi^2(\mathrm{HOG}_{I_n}(\mathbf{p}_s, \mathbf{o}_s) - \mathrm{HOG}_{I^\mathrm{ref}}(\mathbf{p}_s^\mathrm{ref}, \mathbf{o}_s^\mathrm{ref})) \quad (21)$$

is the Chi-square distance of HOG features [10] of a patch at position $\mathbf{p}_s \in \mathbb{R}^2$ of the current image $I_n$ and orientation $\mathbf{o}_s \in \mathbb{R}^2$, where $x_s = \{\mathbf{p}_s, \mathbf{o}_s\}$ and a reference image $I^\mathrm{ref}$ at reference position $\mathbf{p}_s^\mathrm{ref}$ and orientation $\mathbf{o}_s^\mathrm{ref}$. The orientation vector $\mathbf{o}_s$ encodes two aspects: the feature patch rotation (rotation of $\mathbf{o}_s$, i.e. $\mathrm{atan2}(\mathbf{o}_s)$) and feature patch scale (length of $\mathbf{o}_s$, i.e. $\|\mathbf{o}_s\|_2$).

The binary potential $\psi_{s,t}(x_s, x_t)$ is as follows:

$$\psi_{s,t}(\cdot) = \frac{\|\mathbf{p}_t - \mathbf{p}_s - \mathbf{R}_s \mathbf{d}_{st}\|_2^2 + \|\mathbf{p}_s - \mathbf{p}_t - \mathbf{R}_t \mathbf{d}_{ts}\|_2^2}{2 \cdot \|\mathbf{d}_{st}\|_2^2} \quad (22)$$

where $\mathbf{d}_{st(ts)} = \mathbf{p}_{t(s)}^\mathrm{ref} - \mathbf{p}_{s(t)}^\mathrm{ref}$ and $\mathbf{R}_{s(t)} = [o_{x,s(t)}, -o_{y,s(t)}; o_{y,s(t)}, o_{x,s(t)}]$ is a $2 \times 2$ rotation and scale matrix. The proposed binary potential function models the surrounding of each feature point as a weak-perspective model and transforms its neighbor points (with respect to the reference frame) according to a similarity transformation (consisting of translation, rotation, and scaling).

The scalar parameter $\alpha > 0$ is a weighting factor determining the "stiffness" of the feature mesh balancing between feature point independence ($\alpha \to 0$; *i.e.* multi-target tracker) and rigid single object tracking.

### 5.2.2 Tracker Pipeline

A practical application requires some common modifications of the basic tracker pipeline in Sect. 5.2.1. The modifications include an additional *particle resampling* step, where for each frame the initial set of particles are sampled with replacement from the set of particles $\{x_s^{(i)}\}_{i=1,\ldots,p}$ from the previous frame with probability $b_s^N(x_s^{(i)})$. For the tracker to be able to deal with fast moving objects, a *resolution pyramid* approach is applied. The resolution pyramid is only applied to the unary potential function, *i.e.* the feature descriptor is a concatenation of HOG descriptors of patches with the same center position but differing spatial resolution. For each resolution pyramid level (*scale*) the image is downsampled by a factor of 0.5 using bicubic interpolation.

**Slice sampling.** For the slice sampling approach we need to define the boundary functions $A_{\psi_s}(u)$ and $A_{\psi_{s,t}}^{x_t}(u)$. Since $\psi_{s,t}$ is given as an analytic function we can use our automatic inequality solver as described in Sect. 4. An analytic description of the unary potential is not available thus we have to define the boundary manually. We choose to set $A_{\psi_s}(u)$ to the whole image space for $\mathbf{p}_s$, *i.e.* $\mathbf{p}_s \in [1, W] \times [1, H]$, where $W$ and $H$ are the image width and height respectively, and to restrict $\mathbf{o}_s$ to $\mathbf{o}_s \in [-10, 10] \times [-10, 10]$. This way it is ensured that the sampling space is large enough. On the other hand, particles sampled outside the true (sub-)bounds are automatically rejected by the algorithm.

**Metropolis-Hastings sampling.** In order to provide a fair comparison of our slice sampling approach to the state-of-the-art MH-PBP approach, the design of the proposal distribution has to be done very carefully. We propose to use a 4D Gaussian distribution with a covariance matrix $\Sigma$ combined with a suitable coordinate transformation to ensure a well-mixing random walk behaviour. The label space can be divided into two parts, the feature position $\mathbf{p}_s \in \mathbb{R}^2$ and orthogonal feature transformation $\mathbf{o}_s \in \mathbb{R}^2$. The proposal distribution for $\mathbf{p}_s$ is $p(\mathbf{p}_s^{\langle m \rangle} \mid \mathbf{p}_s^{\langle m-1 \rangle}) = \mathcal{N}(\mathbf{p}_s^{\langle m-1 \rangle}, \mathbf{I}_{2\times 2} \cdot \sigma_{xy})$, where $\mathcal{N}(\mu, \Sigma)$ is a Gaussian pdf with mean $\mu$ and covariance $\Sigma$. $\mathbf{I}_{2\times 2}$ is the $2 \times 2$ identity matrix. The vector $\mathbf{o}_s$ is sampled analogously, but in the polar coordinate system with covariance matrix $\Sigma_{\text{polar}} = [\sigma_r^2, 0; 0, \sigma_\phi^2]$, where $\sigma_r^2$ is the variance for the radius and $\sigma_\phi^2$ the variance for the angle. Finally we have to carefully tune the three parameters $\sigma_{xy}$, $\sigma_r$, and $\sigma_\phi$.

### 5.2.3 Tracker Evaluation

**Test sequences.** We use four challenging test sequences (PAPER1, PAPER2, FACEOCC1, and FACEOCC2) to evaluate our proposed method. The self-made PAPER1 and PAPER2 sequences were chosen to challenge the methods on a fast moving deformable object under major scale changes. The sequences have a spatial resolution of $960\,\text{px} \times 540\,\text{px}$ and consist of 563 and 726 frames respectively. The captured object (paper) is textured with patches of similar appearance and shape. The similar appearing features were chosen to stress the relational structure of our tracker model. Thus the only way to distinguish the features is by considering the relative position of the feature patches to each other. The PAPER1 sequence consists of five feature patches with a carefully chosen position pattern which allows unique identification of the features by only having knowledge about the relative distances of the features to each other. The PAPER2 sequence is more challenging since the number of features is increased to 70 and the features are arranged in a grid structure allowing local relational ambiguities. The FACEOCC1 and FACEOCC2 sequences from [2, 5] are designed for evaluating object trackers under major occlusions. The sequences have a spatial resolution of $352\,\text{px} \times 288\,\text{px}$ (FACEOCC1) and $320\,\text{px} \times 240\,\text{px}$ (FACEOCC2) and both consist of 888 frames each. While the FACEOCC1 sequence has only slow object movements, but showing substantial occlusions, the FACEOCC2 sequence challenges with fast movements, illumination changes, object rotation and substantial occlusions. The sequences and tracking results are shown in Fig. 7.

**Parameter selection.** Parameter selection can be split into two parts. The first part consists in MRF model parameter selection. Since the proposed model is relatively robust to changes in $\alpha$, we set $\alpha$ in an ad-hoc fashion for each sequence as follows: $\alpha = 20$ for PAPER1 and PAPER2 and $\alpha = 50$ for FACEOCC1 and FACEOCC2. For the HOG features we set the smallest scale pyramid resolution to $50\,\text{px} \times 50\,\text{px}$. This leads to 3 scales for FACEOCC1 and FACEOCC2 and 4 scales for PAPER1 and PAPER2.

The second part is parameter selection for the PBP framework. We use $N = 20$ PBP iterations and $p = 10$ particles for each node. With this setting both algorithms (MH-PBP and S-PBP) converge well. Since we compare the overall *sampling* behaviour of the proposed method rather than the *belief propagation* convergence behaviour selecting these parameters should be uncritical.

**Evaluation metrics.** We consider the distance $\varepsilon_{\text{track}}$ between the estimated feature position and the groundtruth (manually labeled) position as a quality measure. From this measure we derive two metrics: The rooted mean of squared distances (RMSD) and a quantile box-plot (10%, 25%, 50%, 75%, and 90% quantiles). While the first metric is very sensitive to outliers, the second metric provides more infor-mation about the overall error distribution.

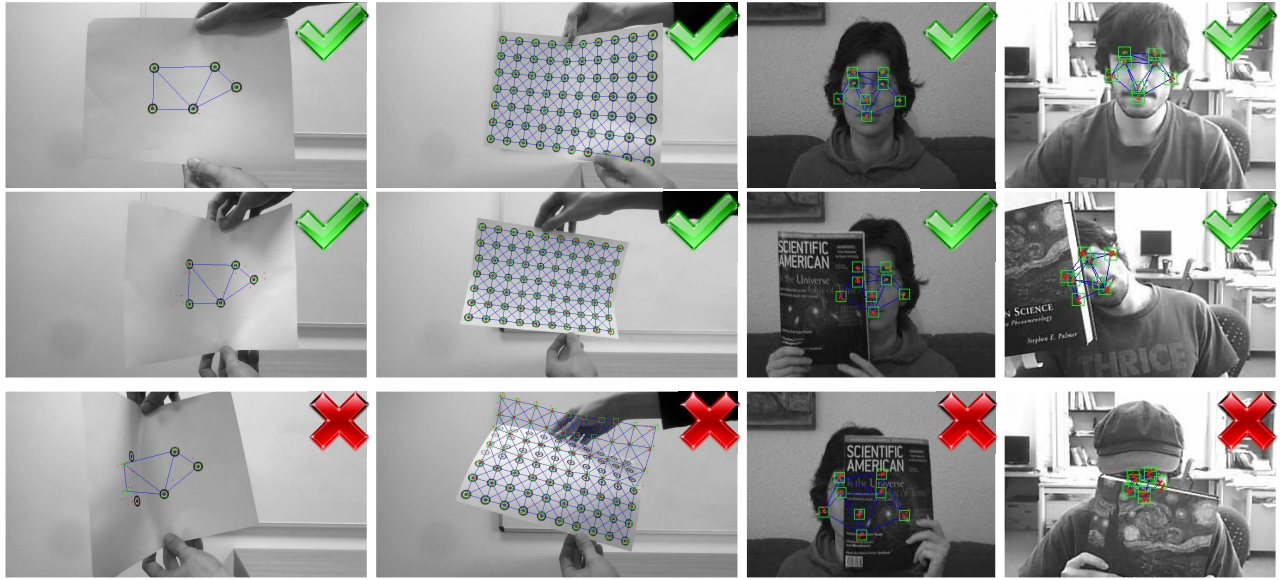

Figure 7. Datasets and tracking results for our proposed method: PAPER1, PAPER2, FACEOCC1, FACEOCC2 (from left to right). First two rows: successful tracking; third row: tracking failure cases.

**Discussion.** The evaluation results comparing S-PBP with MH-PBP using different MCMC iterations are shown in Fig. 8. For MH-PBP, the MH sampling parameters $\{\sigma_{xy}, \sigma_r, \sigma_\phi\}$ are chosen (from the set $\{0.1, 0.2, 0.5, 1.0, 2.0, 5.0\} \times \{0.01, 0.02, 0.05, 0.10, 0.20, 0.50\} \times \{0.01, 0.02, 0.05, 0.10, 0.20, 0.50\}$) such that the RMSD is minimized. Note that for S-PBP such parameter tuning is not necessary. We have evaluated the tracking performance for different MCMC iterations $M = 2$ to 5. The box plots in Fig. 8 show that S-PBP outperforms or performs equally well as MH-PBP for all tested sequences except for sequence PAPER2 with only 2 (and 3) MCMC iterations where both methods fail. This is mainly due to a much higher overall sampling noise of the MH-PBP method compared to S-PBP. We observed that the sampling noise of S-PBP is much less than with MH-PBP at feature positions with high confidence (*i.e.* high belief). On the other hand the sampling noise of S-PBP increases for uncertain feature positions. The RMSD in sequence PAPER2 and FACEOCC1 is higher for S-PBP than for MH-PBP due to temporal tracking failures. These tracking failures are caused by strong local deformations or by occlusions of many feature points. Typical tracking failures are depicted in the bottom row of Fig. 7. It can be observed in such cases that S-PBP leads to much higher tracking error than MH-PBP due to broader particle sampling in uncertain feature positions.

Figure 9 shows an evaluation of MH-PBP under differing (non-optimal) sampling parameters. To this end, we vary each of the three sampling parameters individually and let the other two parameters stay fixed at their optimal values.

Note that the estimation error varies highly, where very high values (usually $> 15\,\text{px}$) indicate a tracking failure. In order to visualize both the performance differences for near-optimal parameters and tracking failures, the error values below and above the $15\,\text{px}$ mark are shown with a differing vertical axis scaling. In Fig. 9, only a comparison for PAPER1 and FACEOCC1 is shown. The other two sequences perform similarly. It can be observed that the tracking performance of MH-PBP strongly depends on careful parameter selection. The parameter $\sigma_{xy}$ has the highest impact on the tracking performance and the optimal parameter value varies strongly between sequences ($\sigma_{xy} = 5$ for PAPER1 and $\sigma_{xy} = 0.5$ for FACEOCC1). Selecting $\sigma_{xy}$ is a compromise between allowing fast object motions and reducing overall localization noise. Selecting $\sigma_r$ and $\sigma_\phi$ has analogous effects on changes in object scaling and rotation. This way one has to incorporate *prior knowledge* about the object motion in order to obtain good tracking results using MH-PBP. Tracked sequences and further comparisons are provided in the supplemental material.

The computational complexity for MH-PBP is $\mathcal{O}(NSpM(1+Vp))$ and for S-PBP is $\mathcal{O}(NSpM(3+2Vp))$ given the number of PBP iterations $N$, nodes $S$, particles $p$, MCMC iterations $M$ and the average number of neighbors per node $V$. This indicates a doubling of computation time of S-PBP compared to MH-PBP which is due to the overhead introduced for computing the interval bounds $A$. A look at the CPU times using fixed parameters for both algorithms ($M = 5, p = 10, N = 20$) verifies this finding: FACEOCC: 0.69 s/frame (S-PBP) vs. 0.33 s/frame (MH-PBP); PAPER2: 7.43 s/frame vs. 3.66 s/frame. Nevertheless we have shown that S-PBP needs significant less MCMC iterations than MH-PBP such that the computational overhead can be typically well compensated.

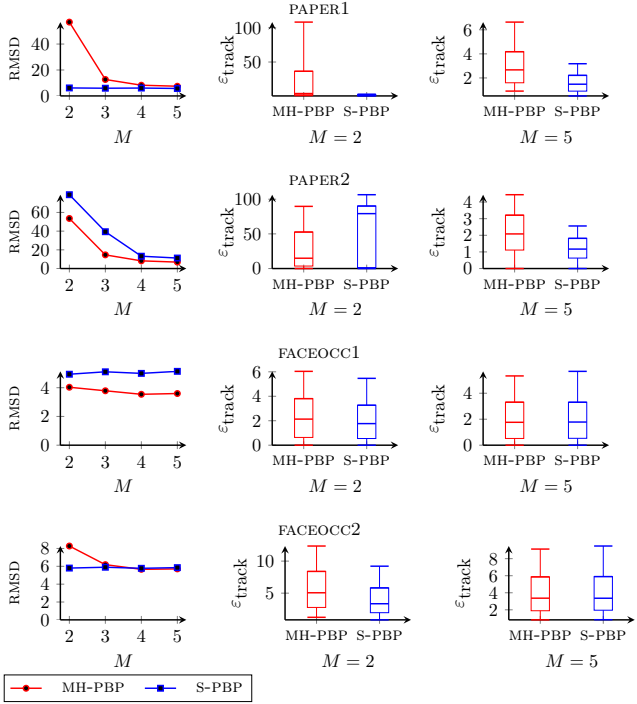

Figure 8. Relational feature tracker evaluation results showing the overal RMSD (for MCMC iterations from 2 to 5) and box plots over the error distance to groundtruth for selected MCMC iterations.

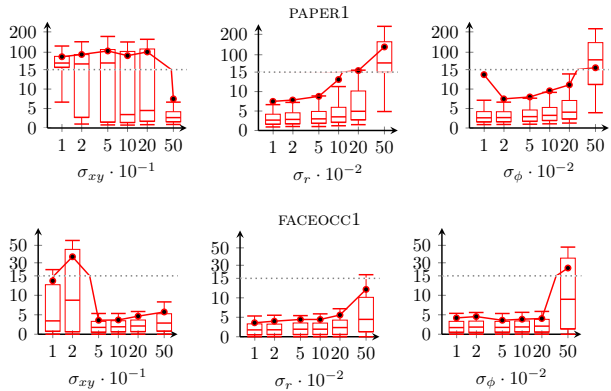

Figure 9. Optimal parameter evaluation for MH-PBP method (with $M = 5$). The vertical axis shows the error distance to groundtruth in px. Note that the vertical axis is stretched for error values lower than 15 px in order to better visualize performance differences.

## 6. Conclusion

We presented a novel particle belief propagation algorithm using slice sampling (S-PBP) instead of Metropolis-Hastings. We exploit the message passing equations to compute the slice sampling bounds, provided the unary and binary potentials are defined by analytic functions or can be bounded by one. We showed on a toy example that S-PBP outperforms MH-PBP in terms of MCMC chain mixing performance. Furthermore we showed that our approach performs equally well or better than MH-PBP on challenging relational feature tracking sequences.

**Acknowledgements** The work is funded by the ERC-Starting Grant (DYNAMIC MINVIP). The authors gratefully acknowledge the support.